\documentclass{article}

    \PassOptionsToPackage{numbers, compress}{natbib}

    \usepackage[preprint]{neurips_2024}

\usepackage[utf8]{inputenc} 
\usepackage[T1]{fontenc}    
\usepackage{hyperref}       
\usepackage{url}            
\usepackage{booktabs}       
\usepackage{amsfonts}       
\usepackage{nicefrac}       
\usepackage{microtype}      
\usepackage{xcolor}         
\usepackage{microtype}
\usepackage{graphicx}
\usepackage{subfigure}
\usepackage{colortbl}

\usepackage{amsmath}
\usepackage{amssymb}
\usepackage{mathtools}
\usepackage{amsthm}
\usepackage{multirow,hhline}
\usepackage{caption}

\usepackage[capitalize,noabbrev]{cleveref}

\newcommand{\tian}[1]{\textcolor{black}{#1}}

\usepackage[textsize=tiny]{todonotes}

\newcommand{\model}[1]{\textit{HydraLoRA}}

\title{\model~: An Asymmetric LoRA Architecture for Efficient Fine-Tuning}

\author{%
  Chunlin Tian\textsuperscript{\dag} \\
  University of Macau\\
  \texttt{yc27402@um.edu.mo} \\
  \And
  Zhan Shi\textsuperscript{\dag} \\
  University of Texas at Austin \\
  \texttt{zshi17@cs.utexas.edu} \\
  \And
  Zhijiang Guo \\
  University of Cambridge \\
  \texttt{zg283@cam.ac.uk} \\
  \AND
  Li Li\textsuperscript{*} \\
  University of Macau \\
  \texttt{llili@um.edu.mo} \\
  \And
  Chengzhong Xu \\
  University of Macau \\
  \texttt{czxu@um.edu.mo} \\
  \And
}

\definecolor{mygray}{gray}{.91}

\begin{document}

\maketitle

\begin{abstract}
Adapting Large Language Models (LLMs) to new tasks through fine-tuning has been made more efficient by the introduction of Parameter-Efficient Fine-Tuning (PEFT) techniques, such as LoRA. However, these methods often underperform compared to full fine-tuning, particularly in scenarios involving complex datasets. This issue becomes even more pronounced in complex domains, highlighting the need for improved PEFT approaches that can achieve better performance. Through a series of experiments, we have uncovered two critical insights that shed light on the training and parameter inefficiency of LoRA. Building on these insights, we have developed \model~, a LoRA framework with an asymmetric structure that eliminates the need for domain expertise. Our experiments demonstrate that \model~ outperforms other PEFT approaches, even those that rely on domain knowledge during the training and inference phases. 
\end{abstract}


\section{Introduction}
Large Language Models (LLMs;~\citep{lm-4,lm-5,lm-6,llama,llama2,chatgpt,gpt4}) are notably powerful, yet their training involves substantial expense. Adapting a single LLM for multiple downstream applications via fine-tuning has emerged as a prevalent method to cater to specific domain needs, balancing performance with practicality. This approach, however, faces a significant challenge due to the extensive memory and computational resources required for full fine-tuning (FFT), i.e., fine-tuning all billions of parameters. A solution to this has been the development of more selective adaptation techniques, involving modifying only a portion of the parameters or integrating external modules designed for new tasks. Key methodologies in this sphere include LoRA~\citep{HuSWALWWC22}, Adaptors~\citep{RebuffiBV17,HoulsbyGJMLGAG19,MahabadiHR21}, and many other variants~\citep{LiL20,LesterAC21, DengWHWGSSXH22, HeZMBN22,ZakenGR22}, all part of what can be generally termed as Parameter-Efficient Fine-tuning (PEFT). PEFT strategies are characterized by freezing the backbone model parameters while only a minimal number of task-specific parameters are introduced and fine-tuned. This method substantially boosts efficiency in the phases of fine-tuning and subsequent deployment, marking a significant advancement in the practical use of LLMs.

\begin{figure}[!t]
    \centering
    \includegraphics[width =0.7\linewidth]{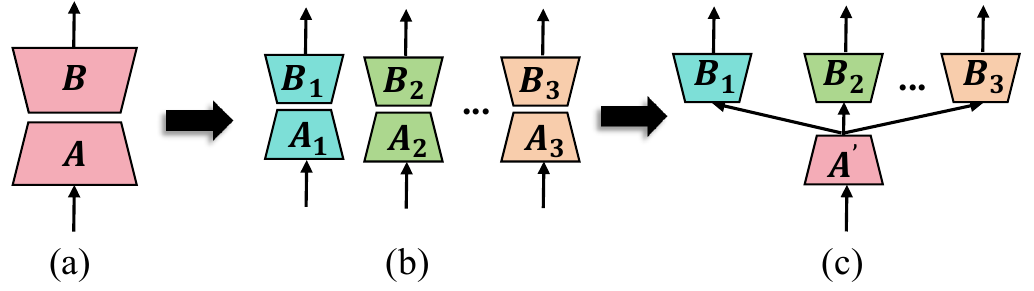}
    \caption{Illustration of LoRA architecture changes in \model~. Only the tunable parameters are shown in this Figure. (a) LoRA architecture with matrix A to achieve low rank and matrix B to recover. (b) under the same parameter count, a monolithic LoRA is split into multiple smaller A and B matrices to avoid training interference. (c) based on (b), \model~ has an asymmetric structure that has a shared A matrix and multiple B matrices.}
    \label{fig:multi-lora-head}
    \vspace{-1.5em}
\end{figure}

While fine-tuning a small subset of parameters offers a streamlined approach for domain adaptation, it's well-recognized that model performance is closely tied to the number of parameters involved~\cite{kaplan2020scaling}. This intrinsic characteristic of methods like LoRA often results in them falling short of the FFT baseline, which updates all parameters, thereby creating a trade-off between efficiency and model quality. This issue of compromised quality in a low-parameter setting becomes even more pronounced in target domains characterized by complex sub-domains and diverse tasks. This situation presents a compelling research question: 

\textit{What is the optimal architecture that can deliver superior model performance while still capitalizing on the efficiency benefits of a reduced parameter footprint?}

\tian{In our research, we carry out a series of exploratory experiments, applying LoRA to the LLaMA2~\citep{llama2} model to adapt it to a new domain encompassing multiple downstream tasks.} As shown in Figure~\ref{fig:multi-lora-head}(a), LoRA adds trainable pairs of rank decomposition matrices A and B in addition to existing weight matrices. Our in-depth analysis of LoRA's mechanics yields several insightful observations and leads to the formulation of key hypotheses. First, rather than employing a single LoRA for the entire domain, it proves more effective to deploy multiple, smaller LoRA heads, each dedicated to a specific downstream task (see Figure~\ref{fig:multi-lora-head}(b)). This suggests that domain or task interference might harmfully impact the training process. We further hypothesize that this interference originates from \textit{``intrinsic components''}---sub-domains or distinct tasks---potentially unknown even to domain experts. Additionally, upon visualizing the parameters of LoRA, we discern a pattern: some parameters predominantly learn the commonalities across all data, while others focus on the unique aspects of each intrinsic component. From these observations, we posit that an optimal LoRA architecture should embody an explicit, asymmetric structure.

Building upon the observations, we propose an improved end-to-end LoRA framework, which we refer to as \model~. From the architecture perspective, unlike LoRA's symmetric structure, \model~ has an asymmetric structure that has a shared A matrix and multiple B matrices (see Figure~\ref{fig:multi-lora-head}(c)). The shared A matrix is used by all samples for parameter efficiency. During the fine-tuning phase, \model~ is designed to autonomously identify ``intrinsic components'' and segregate training samples into distinct B matrices. During the inference phase, \model~ leverages multiple B matrices using Mixture-of-Experts (MoE;~\citep{JacobsJNH91,ShazeerMMDLHD17}) manner. Unlike prior work, \model~ completely eliminates the need for human expertise and assumptions, showing better performance than using domain knowledge to guide the fine-tuning process.

\section{Background and Motivation}
\subsection{LoRA Basics}
LoRA~\cite{HuSWALWWC22} achieves comparable performances to fine-tuning on many benchmarks by freezing the pre-trained model weights $W_{0}$ and inserting trainable rank decomposition matrices into each layer of the pre-trained model. In particular, for each layer, LoRA uses two sequential low-rank matrices $A$ and $B$ to fit the residual weights for adaptation. The forward computation is written as follows:

\begin{equation}
    y\prime=y + \Delta y =W_{0}x + BAx   
    \label{eq:lora}
\end{equation}

where $y \in R~\textsuperscript{d}$ is the output and the $x\in R~\textsuperscript{k}$ denotes the input. $B \in R~\textsuperscript{d×r}, A \in R~\textsuperscript{r×k}$ with $r \ll min(d, k)$. Normally matrix $B$ is initialized with zeroes and matrix $A$ is initialized with Kaiming Uniform~\cite{he2015delving} to force $\Delta~y = 0$ at the beginning. 

\subsection{LoRA's Practical Dilemma}
\begin{figure}[!ht]
    \centering
    \begin{minipage}{0.51\textwidth}
        \centering
\includegraphics[width=0.65\linewidth]{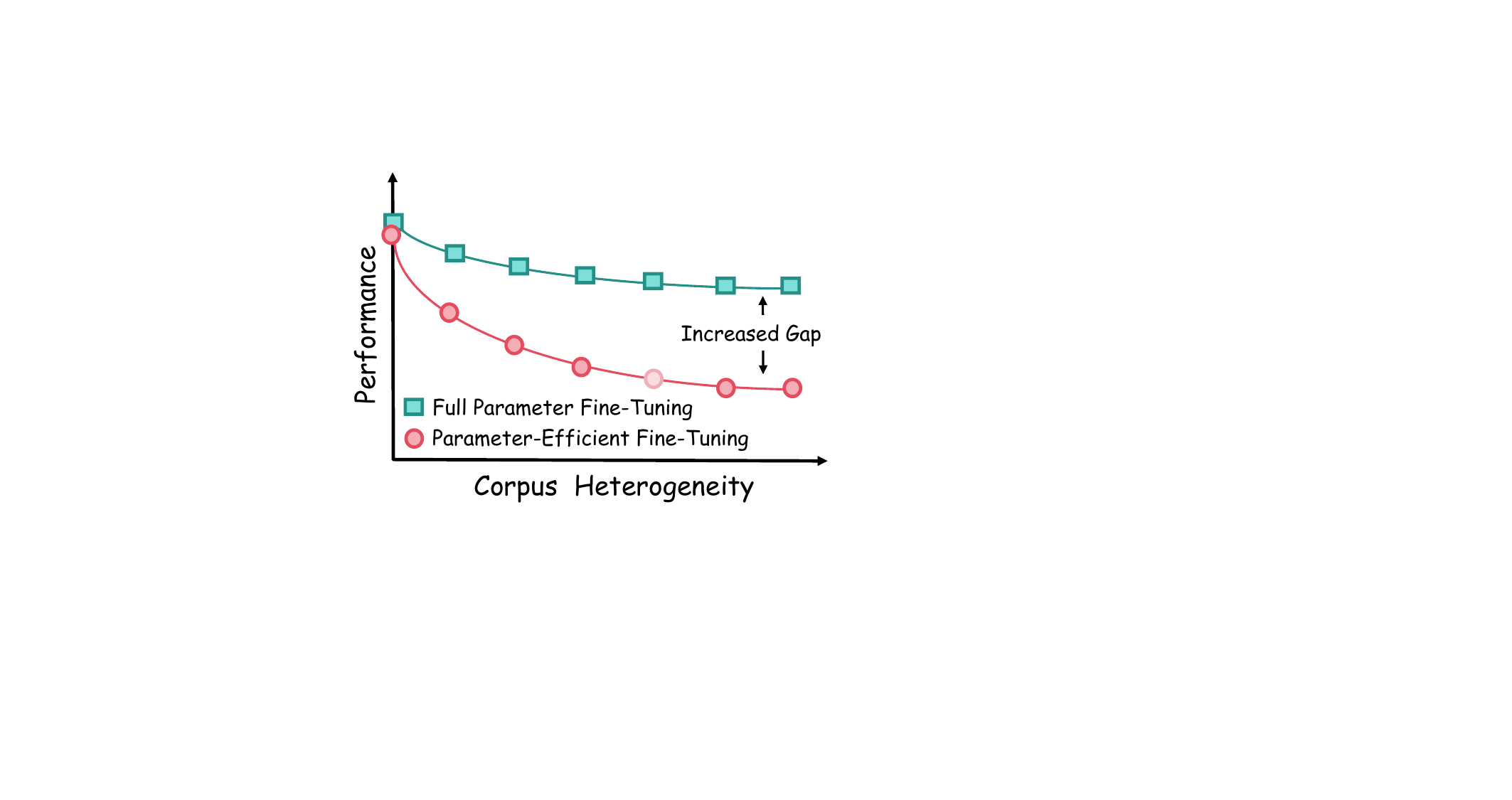}
        \vspace{-0.5em}
        \caption{\small Performance impact of corpus heterogeneity on full fine-tuning vs. parameter-efficient fine-tuning. Heterogeneity signifies the diversity within the dataset, often leading to interference due to its varied content and style~\cite{S-LoRA-FL}. Parameter-efficient approaches are particularly sensitive, suffering greater performance losses in heterogeneous cases.}
        \label{fig:Heterogeneity}
    \end{minipage}\hfill
    \begin{minipage}{0.46\textwidth}
        \centering
        \captionof{table}{\small Performance on instruction tuning with Dolly-15K \cite{DollyV2} and evaluated with MMLU \cite{mmlu} with different ranks. For LoRA (Split) decomposes high-rank LoRA modules into smaller, equivalent low-rank components ($r \times n$). $n$ is  the number of LoRAs, $r$ denotes the rank of each LoRA.}
        \resizebox{0.9\linewidth}{!}{
        \begin{tabular}{c|c|c|c}
        \bottomrule[1.5pt]
        \rowcolor{mygray}\textbf{Schemes}  &  \textbf{$r \times n$}  &  \textbf{MMLU} $\uparrow$ & \% \textbf{Parameter} \\
        \toprule[0.75pt]
        LoRA & $8 \times 1$ & 43.22 & 0.062\\
        LoRA & $16 \times 1$ & 45.45 & 0.124 \\
        LoRA & $32 \times 1$ & 46.59 & \textbf{0.248}\\ \midrule
        LoRA (Split) & $16 \times 2$ &46.82 & 0.248\\
        LoRA (Split)& $8 \times 4$ &\textbf{46.94} & 0.248\\
        LoRA (Split)& $4 \times 8$ &46.83 & 0.248\\
        \bottomrule[1.5pt]
        \end{tabular}}
        \label{tab:split_lora}
    \end{minipage}
    \vspace{-1em}
\end{figure}

Parameter count has a clear impact on the performance of neural models~\cite{kaplan2020scaling,gpt4}. Yet, Parameter-Efficient Fine-tuning (PEFT) methods, such as Adapter~\citep{HoulsbyGJMLGAG19} and prefix-tuning~\citep {LiL20}, focus on fine-tuning a limited set of parameters. These approaches present a practical dilemma: while restricting the number of tuned parameters is essential for training efficiency, it hinders the model's ability to learn from diverse datasets. This trade-off becomes particularly evident when considering corpus heterogeneity~\cite{S-LoRA-FL}. Figure~\ref{fig:Heterogeneity} reveals a notable performance disparity between PEFT techniques and full fine-tuning (FFT), with the gap widening in scenarios involving a more diverse or heterogeneous training corpus.

\subsection{Observations}
\label{section: Observation}
In this work, we aim for a PEFT approach that strikes a better balance between maximizing the learning capability for heterogeneous data and minimizing the number of parameters involved. A key goal is to ensure that our enhanced technique exhibits robust generalization across unseen tasks, independent of any prior task-specific knowledge. 
To achieve our objectives, we focus on LoRA and conduct a series of experiments as Table \ref{tab:split_lora} to gain a deeper understanding of its mechanisms. Our methodology involves leveraging data from diverse tasks within a domain, and training distinct LoRA heads for each domain, leading to our first observation:


\textit{\textbf{Observation I}: With the same parameter count, rather than employing a single LoRA for the entire domain dataset, it proves more effective to deploy multiple, smaller LoRA heads, each dedicated to a specific downstream task.}

This suggests that interference among tasks might harmfully impact the training process. Furthermore, we posit that this interference is NOT exclusive to this explicit multi-task training. This interference could happen in any training setting since all datasets inherently consist of multiple implicit \textit{intrinsic components}, such as sub-domains or tasks within a domain that is even unknown to domain experts.
To better understand how multiple LoRA heads mitigate the interference among intrinsic components, in Figure~\ref{fig:LoRA_modules}, we employ the t-SNE technique~\cite{t-SNE} to visualize the parameters of matrix A and B across all heads. This analysis yields another critical observation:

\begin{figure}[!t]
    \centering
    \includegraphics[width =0.33\linewidth]{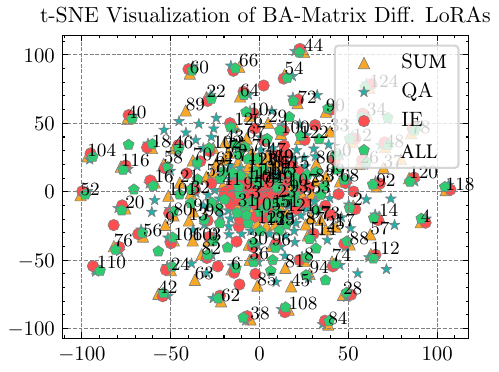}
    \includegraphics[width=0.325\linewidth]{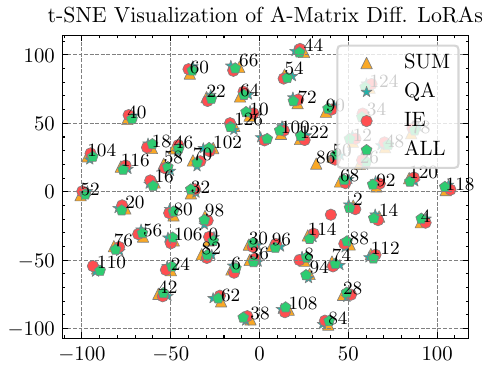}
     \includegraphics[width=0.32\linewidth]{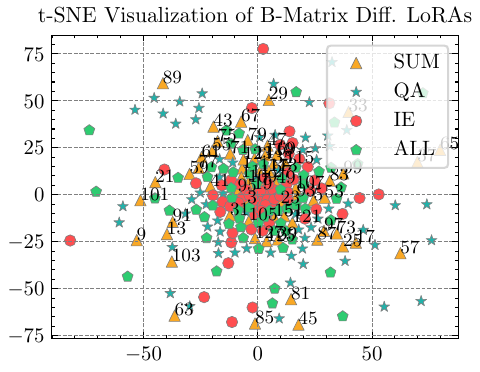} 
     \caption{\small Breakdown analysis of LoRA modules.
     Compare fine-tuned LoRA modules of Dolly-15K~\cite{DollyV2} with three subtasks of Dolly-15K including ``\textit{summarization (Sum)}'', ``\textit{closed QA (QA)}'' and ``\textit{information extraction (IE)}''  using t-SNE.
     Consider LLaMA2-7B (random seed=42), which contains 32 decoder layers, corresponding to 32 adaptive modules. Each module consists of \{\textbf{0}: q\_proj of A, \textbf{1}: q\_proj of B, \textbf{2}: v\_proj of A, \textbf{3}: v\_proj of B\} submodules. This makes a total of $32\times4$ submodules.  Left displays all submodules. Center shows all even submodules, i.e. the A matrix. Right represents all odd submodules, i.e. the B matrix. It can be seen that the differences in the fine-tuned LoRA modules for different tasks arise mainly from the B matrix. }
    \label{fig:LoRA_modules}
    \vspace{-1em}
\end{figure}

\textit{\textbf{Observation II}: When multiple LoRA heads are trained individually on different data, the parameters of matrix A from different heads tend to converge, while those of matrix B are distinguishable.}

In detail, the parameters of matrix A across all heads exhibit a high degree of similarity, leading to their overlaps in the figure. Conversely, the parameters of matrix B from different heads are distinct and easily distinguishable. We posit that this divergence is an artifact of the initialization schemes, with matrix A inclined toward capturing commonalities across domains, while matrix B adapts to domain-specific diversities. The distinction between matrix A and B offers valuable insights for enhancing both parameter efficiency and effectiveness. From an efficiency standpoint, our hypothesis suggests that the parameters of matrix A could potentially be shared across multiple heads, thereby reducing redundancy.
Regarding effectiveness, since the parameters of matrix B of different heads are dispersed, suggesting that using a single head to adapt to multiple domains might be less effective than using individual heads for each domain, which minimizes the interference between domains.

Building upon our observations, we propose an optimized LoRA architecture designed to enhance cost-effectiveness. In this architecture, we share the parameters of A matrix across various sub-domains or tasks to improve parameter efficiency, while deploying multiple B matrices, each tailored to handle different intrinsic components. This design allows for a more effective adaptation to the specific characteristics of each component. While these intrinsic components can be manually identified using prior knowledge of the training data, we also introduce end-to-end methods using Mixture-of-Experts (MoEs)~\cite{MoE}, which will be detailed in the methodology section. This automatic approach facilitates flexibility and applicability, particularly in scenarios where prior knowledge is limited or unavailable.

\section{\model~}
\label{section:design}
\begin{figure}[!t]
    \centering
    \includegraphics[width =0.9\linewidth]{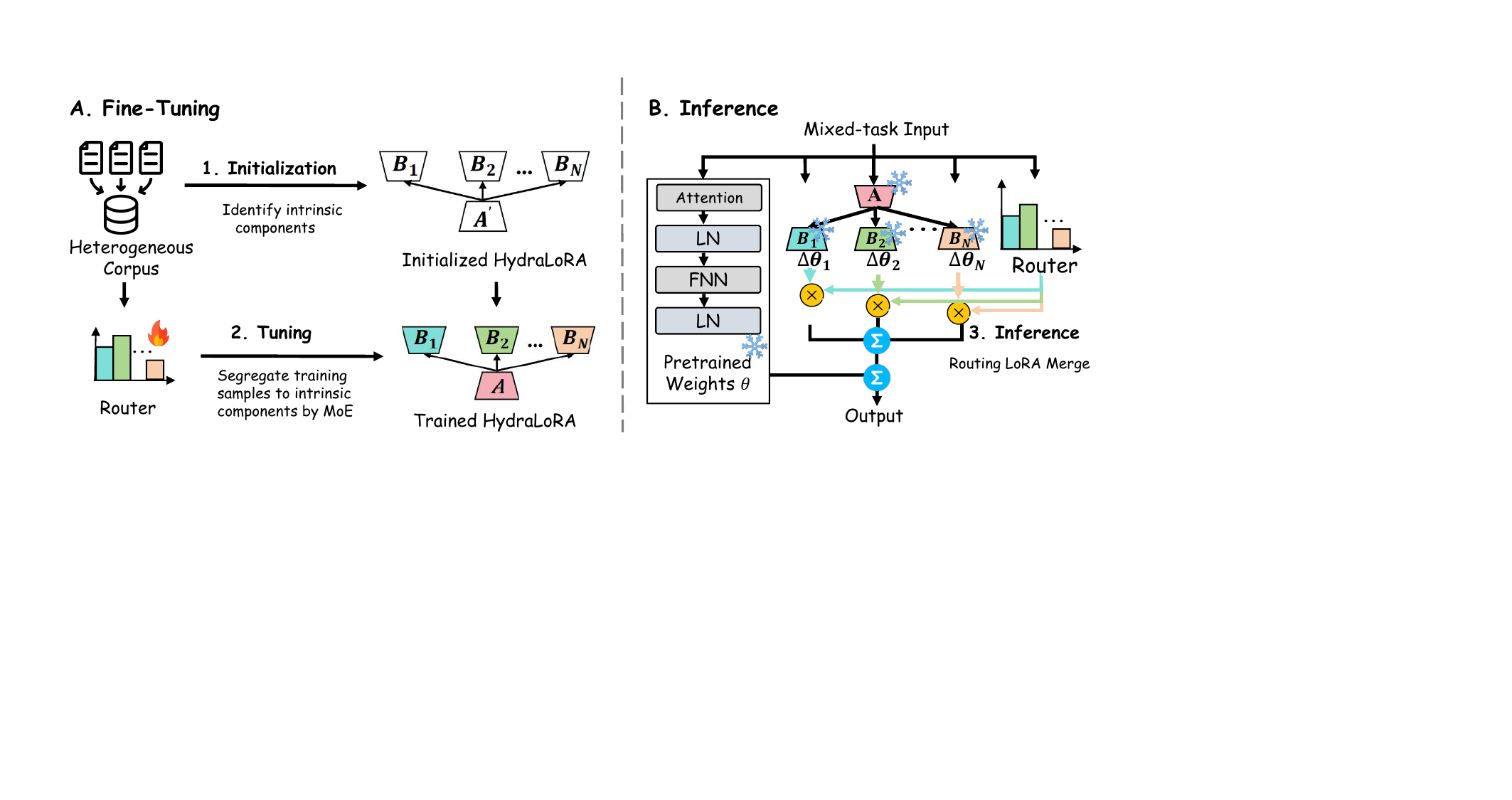}
    \caption{
    \tian{\small Architecture and workflow of \model~. During the fine-tuning stage, \model~ first adaptively identifies and initializes $k$ of intrinsic components without specific domain knowledge. It then employs a trainable MoE router that treats each intrinsic component as an expert to automatically segregate training samples into intrinsic components for fine-tuning.
    During the inference stage, \model~ merges multiple $B$ matrices flexibly and dynamically through a trained router.}}
    \vspace{-1em}
    \label{fig:framework}
\end{figure}

In this section, we introduce the proposed \model~, an asymmetric LoRA architecture for
efficient fine-tuning, as illustrated in Figure \ref{fig:multi-lora-head}. After that, we show the workflow of \model~ as Figure~\ref{fig:framework}.

\subsection{Asymmetric LoRA architecture}
\tian{
The LoRA method updates two low-rank matrices $A$ and $B$, and uses $AB$ as the change of a pretrained and frozen weight $W_{0}$ of a linear layer as shown in Eq. \ref{eq:lora}. The integral parameters are fine-tuned for the whole corpus in the original LoRA, which causes difficulty in learning the various knowledge aspects.
Drawing from a detailed breakdown analysis of LoRA, a potential solution is to segment the entire LoRA into ``Hydra'' structured LoRA variants, that is, characterized by a central shared matrix $A$ and several distinct matrices $B$, fostering a blend of shared knowledge and specialized functionalities. As Figure \ref{fig:multi-lora-head}, \model~ is to fine-tune LoRAs to achieve robust performance without redundancy, thereby benefiting the entire heterogeneous corpus. The asymmetric LoRA architecture can be formulated as: }
\begin{equation}
\begin{aligned}
\mathit{W} & = \mathit{W_{0}} +  \Delta \mathit{W} \\
                & = \mathit{W_{0}} +  \sum_{i=1}^{N} \omega_{i} \cdot B_i A 
\end{aligned}
\end{equation}
The matrics $B_i \in \mathbb{R} ^{d \times r}$ and shared $A \in \mathbb{R}^{r \times k}$. The hyper-parameter $N$ denotes the number of $B$ matrices. The term $\omega_{i}$ modulates these contribution weights for head $B_{i}$.

\subsection{Workflow of \model~}
Figure~\ref{fig:framework} illustrates the workflow of \model~. Initially, \model~ delves into the adaptive identification and initialization of LoRA modules within a heterogeneous corpus, aligning them with task relevance through the application of $k$-means \tian{or developer-specified size}. Subsequently, we propose a Mixture-of-Experts (MoE) framework that handles $B$ matrices as expert adapters to ensure computational efficiency throughout the fine-tuning (Section~\ref{sec:train}) and inference (Section~\ref{sec:inference}) stages by freezing the rest of the LLM parameters. During inference, it flexibly and dynamically merges multiple $B$ matrices through the MoE router.

\subsubsection{Fine-tuning}
\label{sec:train}
Motivated by Mixture-of-Experts (MoE;~\cite{JacobsJNH91,ShazeerMMDLHD17}), where experts are selectively activated by a gating mechanism (Router) in response to different inputs.  In \model~, we substitute each expert with a lightweight LoRA adapter. During fine-tuning, while weights of LLMs remain frozen, the experts and router layers are trained from scratch. In order to achieve a unified approach to the distinct forward processes of multiple $B$ matrices, we define a set of experts, denoted as $(E_{1}, \ldots, E_{N})$, to learn the updated matrix $\Delta W$. As \model~ fine-tunes the experts using the heterogeneous corpus, the shared matrix $A$ inherently captures collaborative knowledge to augment intra-gains, and different matrices $B$ foster knowledge modularity to mitigate fine-tuning inter-offsets. Based on this structure, the forward process of \model~ is expressed as:
\begin{equation}
    y  =\mathit{W}_0 x +\sum_{i=1}^N \omega_{i}  E_i A x  \quad (MoE)
\end{equation}

where $N$ denotes the number of experts, i.e., $B$ matrices.To regulate these contributions, we introduce a gate function (router network) commonly consisting of a dense layer with trainable weights (transformation matrix) $\mathit{W}_{g} \in \mathbb{R} ^{r \times N}$ followed by a softmax function which takes an intermediate token representation $x$ as input and combines the output of each expert based on the gating scores $(\omega_1, \dots, \omega_N)$:
\begin{equation}
    \omega_{i} = \text{softmax}(\mathit{W}_{g}^{T} x) \quad (Router)
\end{equation}

\subsubsection{Inference}
\label{sec:inference}
\tian{
During inference, \model~ merges adapters by enabling routing computation based on the input. Specifically, since matrices B operate as linear functions, we initially compute a weighted average of the experts. Following this, we apply a PEFT transformation using the combined expertise. 
The \model~ significantly enhances training efficiency through an extremely parameter-efficient MoE formulation. Additionally, the intrinsic structural modularity of \model~ facilitates rapid recovery and merging of the trained parameters during inference, leading to substantial memory savings.}

\begin{table}[!t]
\small
\centering
\caption{\small Comparative performance of different tuning schemes across multiple benchmarks on a single domain. 8-shot for GSM8K, zero-shot for others. \#$\Bar{B}$ refers to the average $B$ matrix number.}
\resizebox{0.9\linewidth}{!}{
\begin{tabular}{c|cccccc|c|cc}
\bottomrule[1.5pt]
\rowcolor{mygray} &  &  & & \multicolumn{2}{c}{\textbf{HumanEval}} & &  &  &  \\
\rowcolor{mygray}
 \multirow{-2}{*}{\textbf{Schemes}}& \multirow{-2}{*}{\textbf{MMLU}}  &\multirow{-2}{*}{\textbf{Medical}}  & \multirow{-2}{*}{\textbf{Law}} & P@1&P@10&\multirow{-2}{*}{\textbf{GSM8K}}  & \multirow{-2}{*}{\%\textbf{Param}} & \multirow{-2}{*}{\#\textbf{A}}  & \multirow{-2}{*}{\#$\Bar{\textbf{B}}$} \\
 \toprule[0.75pt]
LLaMA2-7B \citep{llama2} & 38.88 &  35.98 & 33.51 & 13.10 & 20.34 & 10.38 & - & -& -\\
Full Fine-Tuning & 49.91 & 46.78&46.08 &20.24 &32.93& 25.70&100 & -& -\\ \midrule[0.75pt]
Prompt Tuning \citep{LesterAC21} & 39.91 & 37.59 &35.02 & 13.66& 21.55 &13.18 &  0.001& -& - \\
P-Tuning$_{(256)}$  \citep{P-tuning}  & 41.11 & 39.81 &36.72 & 13.60&21.13 &15.56& 0.193& -& -\\
Prefix Tuning \citep{LiL20} & 41.78 & 40.28& 36.54 & 13.23& 22.56&16.89 & 0.077& -& -\\
(IA)$^3$ \citep{IA3} & 40.45 & 37.12 & 35.25&13.54 &23.17& 13.98&0.009& -& -\\
AdaLoRA$_{(r =8)}$ \citep{Adalora}&  44.32& 42.83 & 39.36& 14.81 &23.78 &19.51& 0.093 & 1& 1\\
\midrule  
LoRA$_{(r =8)}$ \citep{HuSWALWWC22} & 43.22 & 41.59 & 37.85& 15.67&22.95&18.24 & 0.062& 1& 1\\
LoRA$_{(r =16)}$ & 45.45 & 43.10 & 39.64 &16.71 &25.60 & 20.32& 0.124 & 1&1\\
LoRA$_{(r =32)}$ & 46.59& 44.32 & 40.81 &17.12 & 25.89&20.67 & 0.248 & 1&1\\
LoRA-Split$_{(4 \times 8)}$ & 46.94 & 45.28 & 41.35& 18.20 & 26.85 & 21.92 &  0.248 & 4& 4\\
 \midrule[0.75pt]
 \textbf{\model~}$_{(r=8)}$  & \textbf{47.22}&\textbf{45.71} &\textbf{42.18}& \textbf{18.31}&\textbf{27.43}& \textbf{22.27}& \textbf{0.124}& 1& 3\\ 
\toprule[1.5pt]
\end{tabular}}
\vspace{-1em}
\label{table:task-corpus}
\end{table}

\section{Experiments}
In this section, we detail the principal experiments. We begin with an overview of the experimental setup and implementation intricacies. Following this, we share our findings and offer a succinct interpretation.

\subsection{Experiment Setting}
\paragraph{Dataset and Benchmarks} To explore the properties and commonalities of the LoRA asymmetric structure, we conduct experiments on both single and multiple domains to evaluate the effectiveness of \model~ for profiling intrinsic components.
$\bullet$ Single domain. 1) \textit{General}: we fine-tune with the general instruction tuning databricks-dolly-15k~\cite{DollyV2} for generic language capability and evaluate with MMLU~\cite{mmlu}.
2) \textit{Medical}: we fine-tune with GenMedGPT and clinic-10k from ChatDoctor~\cite{chatdoctor} for medicine applications and evaluate medical tasks in MMLU.
3)  \textit{Law}: we fine-tune with two legal instruction tuning datasets Lawyer-Instruct~\cite{Lawyer-Instruct} and US-Terms~\cite{legal_lama} then evaluate with law tasks in MMLU.
4)  \textit{Math}: we fine-tune with the training split of GSM8K~\cite{gsm8k} for mathematical reasoning and evaluate with test set of GSM8K.
5)  \textit{Code}: we fine-tune with CodeAlpaca~\cite{codealpaca} for code generation and evaluate with HumanEval~\cite{Humaneval}.        
$\bullet$ Multi-task domain. We select a portion of the Flanv2~\cite{flanv2} datasets covering Natural Language Understanding (NLU) and Natural Language Generation (NLG), which can be grouped into 10 distinct task clusters. Then we evaluate it with the Big-Bench Hard (BBH) \citep{bbh} benchmark. A detailed description of the benchmarks can be found in Appendix \ref{appendix:datasets}.

\paragraph{Baselines} 
$\bullet$ First, we compare \model~ with different PEFT methods on single datasets: 1) \textit{Full fine-tuning};
2) \textit{Prompt Tuning}~\citep{LesterAC21};
3)  \textit{P-Tuning}~\citep{P-tuning};
4) \textit{Prefix Tuning}~\citep{LiL20};
5) \textit{IA$^{3}$}~\citep{IA3};
6) \textit{AdaLoRA}~\citep{Adalora}. $\bullet$ Second, we extend the experiments exploring \model~ on multiple datasets compared with more weighted average methods: 1) 
Lorahub \cite{HuangLoraHub2023} employs black-box optimization to learn weights of 20 randomly selected LoRAs for new tasks, using weighted averaging without needing gradient calculations.
2) LoRA MoE \cite{Zadouri2023} combines lightweight experts (LoRA) with MoE architecture for high efficiency, generalizing to new tasks without prior knowledge.
A detailed description of the baseline models can be found in Appendix \ref{app_baselines}.

\subsection{Overall Performance }
The experimental results of \model~ and the competing baselines are presented in Table~\ref{table:task-corpus} with a single domain and Table~\ref{table:mix-task-corpus} with the mixed task domain. 
The evaluation of diverse tasks demonstrates that \model~ consistently outperforms all other schemes. The performances rooted in LoRA outperform those of conventional PEFT methodologies. Compared to the default single LoRA configuration (rank=8), the Hydra architecture, enriched by the integration of several B matrices, effectively addresses the inherent conflicts among intrinsic components of the corpus. Furthermore, with equivalent parameters (rank=16), the model shows superior performance, confirming the effectiveness of the adopted parameters. Based on Table~\ref{table:task-corpus} and Table~\ref{table:mix-task-corpus}, we propose three research questions that confirm the aforementioned observations.

\begin{table}[!t]
    \centering
    \caption{\small Comparative performance of different tuning schemes, including base model (Base), LoRA tuning (LoRA), LoraHub learning, multi-LoRA tuning with MoE inference (LoRA MoE) and our proposed \model~ learning across mix-task domain on the BBH benchmark with LLaMA2-7B, LLaMA2-13B as the base LLM (3-shot). Refer to Appendix \ref{app:results} for details.}
    \resizebox{0.9\linewidth}{!}{
\begin{tabular}{c|c|ccccc}
\bottomrule[1.5pt]
\rowcolor{mygray}\multicolumn{2}{c|}{\textbf{Metrics}}  & \textbf{Base} \citep{llama2}  & \textbf{LoRA} \citep{HuSWALWWC22} & \textbf{Lorahub} \citep{HuangLoraHub2023}  & \textbf{LoRA MoE} \citep{Zadouri2023} &  \textbf{\model~} \\
\toprule[0.75pt]
\multirow{2}{*}{Performance} & 7B & 31.6 & 36.8 & 39.7 & 40.3 & \textbf{41.5} \\ 
 & 13B & 38.4 & 40.1 & 41.9 & 43.7 & \textbf{44.1} \\ \midrule[0.75pt]
\multicolumn{2}{c|}{\# of A/B for training} & 0/0 & 1/1 & 48/48 & 48/48 & 1/10 \\
\multicolumn{2}{c|}{\# of A/B for inference} & 0/0 & 1/1 & 20/20 & 48/48 & 1/10 \\
\multicolumn{2}{c|}{\% Params} & - & 0.062 &  1.240 &  2.976 & 0.341 \\
\toprule[1.5pt]
\end{tabular}}
\vspace{-1em}
\label{table:mix-task-corpus}
\end{table}

\vspace{-0.5em}
\paragraph{RQ1: Is it more effective to use multiple smaller LoRA heads for specific tasks rather than one single LoRA for the entire domain dataset, given the same parameter count?}
Comparing the high-dimensional LoRA configuration with \( r = 32 \)  against a segmented version using LoRA-Split, a variant introduced by HydraLoRA, which divides the model into four distinct components each with  \( r = 8 \). That is, multiple vanilla LoRAs are directly utilized to capture the differences between data. We observe a noteworthy trend in the performance across a variety of tasks as detailed in Table~\ref{table:task-corpus}. It illustrates the superior performance of LoRA-Split in comparison to the traditional LoRA approach, across all the evaluated scenarios. This enhancement in performance is a strong indication of the detrimental impact that task interference can have on the training process. By segregating the tasks into discrete components, LoRA-Split effectively minimizes the conflict and interference between tasks, thereby promoting a more efficient and focused environment.

The concept of LoRA-Split hinges on the construction of different intrinsic component compositions, employing LoRA as a foundational technique to strategically mitigate the interference conflict. This architectural innovation has proven to be a pivotal factor in enhancing model performance. However, it's important to note that while LoRA-Split marks a significant advancement in model efficiency and task handling, it also introduces a certain level of parameter redundancy. The segmented approach of LoRA-Split inevitably leads to an increase in the overall model parameters, which can be manifold in comparison to the traditional, singular LoRA model. This increase in parameters, while contributing to the model's robustness and capability to handle multiple tasks simultaneously, also poses new challenges in terms of computational resources and model optimization.

\paragraph{RQ2: Will multiple LoRA heads, individually trained on different data, improve efficiency by distinguishing matrix B parameters?}
We evaluated the Hydra structure LoRA --- \model~ that is characterized by a shared LoRA A matrix, while maintaining distinct B matrices that are trained separately. This configuration was meticulously compared with both the standard LoRA and the LoRA-Split approaches, emphasizing efficiency parameters.

According to the results presented in Table~\ref{table:task-corpus}, unlike split which straightforwardly adopts multiple vanilla LoRAs, \model~ adopts an asymmetric LoRA structure that not only improves parameter efficiency by separating the uses of A matrix for commonalities and B matrices for diversities with a notably smaller adapter parameter set, but also employs a trainable router to improve the composition of multiple B matrices that outperforms the LoRA-Split approach. 
This finding is significant as it suggests that \model~ not only enhances performance efficiency but also boosts overall system effectiveness. This may be driven by 1) different B matrices capturing different features of the data-intrinsic knowledge, mitigating mutual interferences, and avoiding performance offsets. 2) Module A maintains the collaborative knowledge by taking the strengths of each and integrating them to improve the model performance.

\paragraph{RQ3: How does \model~ fare against other merge methods in complex, multi-task domains, considering scalability and robustness?}
While we hypothesize that the asymmetry is mainly rooted in the different initialization methods of A and B matrices, it is possible that this behavior varies on different model architectures and datasets. To the best of our ability, we extended the experiments exploring \model~ on multiple datasets. LoRA MoE and their variants typically aim at tackling multi-tasks by employing multiple independent LoRAs. This makes them suitable for handling various domains. However, for a single dataset like ours, a ``default'' MoE method might not be optimal. HydraLoRA addresses this by constructing asymmetric structures and utilizing multiple B matrices to capture the specific nuances within the single dataset. The effectiveness of this approach is demonstrated by the experimental results in Table \ref{table:mix-task-corpus}.


\begin{figure}[!t]
    \centering
    \includegraphics[width=0.64\linewidth]{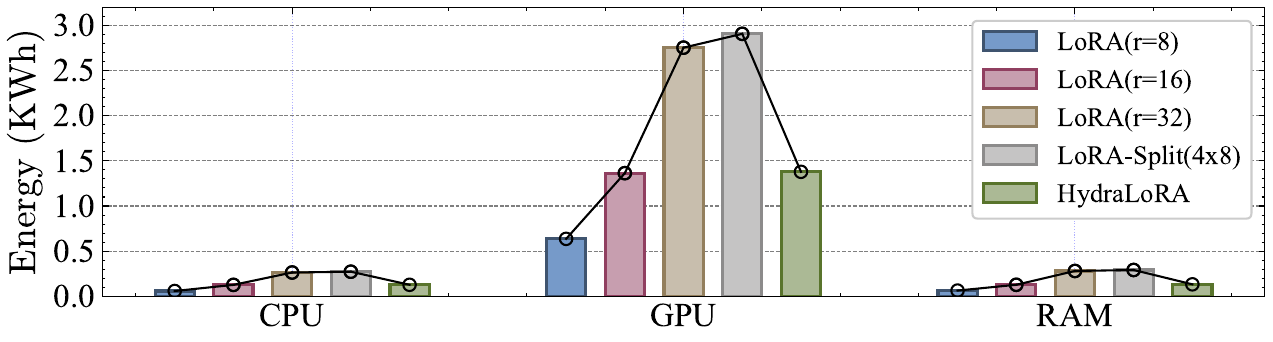}
     \includegraphics[width=0.35\linewidth]{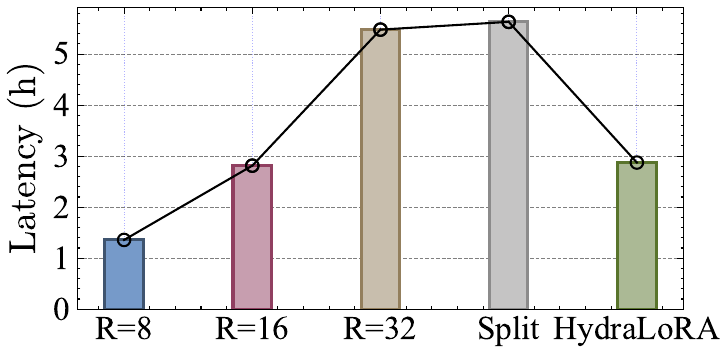}
     \vspace{-1em}
    \caption{Energy consumption and latency during fine-tuning with different LoRA approaches (fine-tuning LLaMA2-7B with GSM-8K).}
    \vspace{-1.5em}
    \label{fig:Energy}
\end{figure}

\subsection{Energy and Throughput Analysis}
\paragraph{RQ4: How does the ``Hydra'' structure in \model~ enhance system efficiency, particularly in reducing training energy consumption and latency?}
We evaluate the system efficiency of \model~  from two perspectives: training energy consumption and latency.
The following experiments were executed on a GPU infrastructure consisting of 4 NVIDIA A40 GPUs and a CPU powered by an Intel(R) Xeon(R) Gold 6330 CPU clocked at 2.00GHz. Power consumption measurements were recorded using CodeCarbon~\cite{carbon}. Figure~\ref{fig:Energy} shows the results of various fine-tuning approaches for GSM-8K using the LLaMA2-7B model. we can see that \model~ effectively speeds up the training process $1.96 \times$ and reduces 49.6\% energy cost compared to LoRA (rank=32). While the energy consumption and latency of LoRA-Split exceeds the LoRA (rank=32). This is for the reason that \model~ jointly considers inherent knowledge modularity and collaboration, which utilizes the ``Hydra'' structure with a shared A matrix and different B matrix. In this way, it only employs rank=16 training overhead but expands to a performance enhancement of more than rank=32. Overall, this experiment demonstrates the parameter effectiveness of \model~.

\begin{figure}[!t]
    \centering
    \includegraphics[width= 0.7\linewidth]{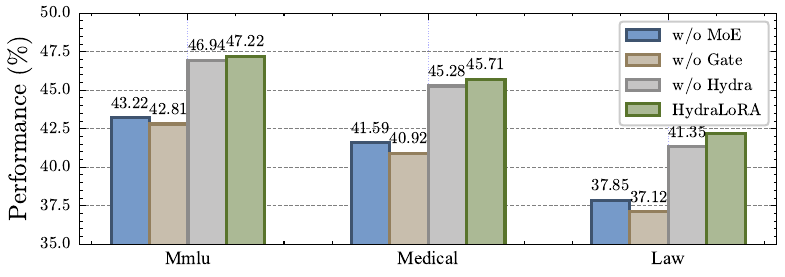}
    \vspace{-0.5em}
    \caption{Comparative performance of ablation study for \model~ across multiple benchmarks.}
    \label{fig:ablation}
    \vspace{-1em}
\end{figure}

\subsection{Ablation Study}
\paragraph{RQ6: What impact do the MoE architecture and the gate function have on the fine-tuning process?}
To delve deeper into understanding the contributions of each component in \model~. we present the results of our ablation study in Figure \ref{fig:ablation}. The variant \textit{w/o} MoE (essentially reverts to LoRA)  excludes the MoE architecture. Similarly, the \textit{w/o} gate variant employs uniform expert weights bypassing the gate function. The \textit{w/o} hydra adopts multiple vanilla LoRAs in a straightforward way. Figure \ref{fig:ablation} indicates that the full \model~ model outperforms its variants, showing that both the MoE architecture and gate function significantly contribute to its effectiveness across various language understanding domains.

\subsection{ Hyper-parameter Analysis}
\begin{figure}[!t]
    \centering
    \begin{minipage}[c]{0.58\textwidth}
        \centering
        \includegraphics[width= 0.95\linewidth]{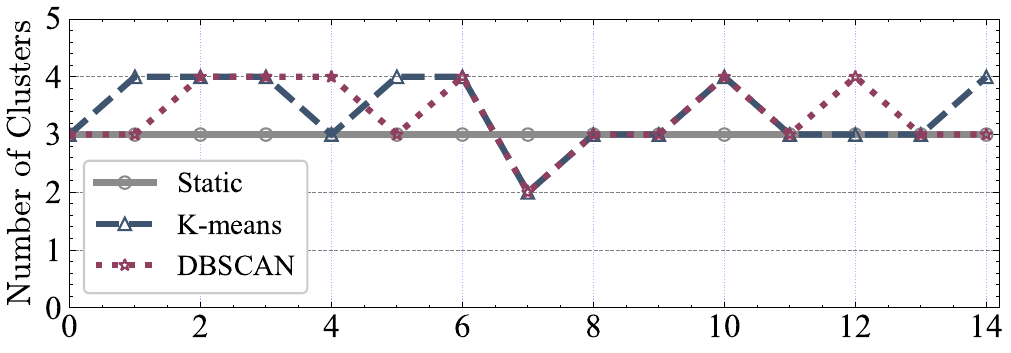}
        \caption{\small Number of clusters generated by different approaches including developer-specific (static), k-means, and DBSCAN.}
        \label{fig:k-value_gen}
    \end{minipage}%
    \hspace{5mm}
    \begin{minipage}[c]{0.37\textwidth}
        \centering
        \includegraphics[width= 0.95\linewidth]{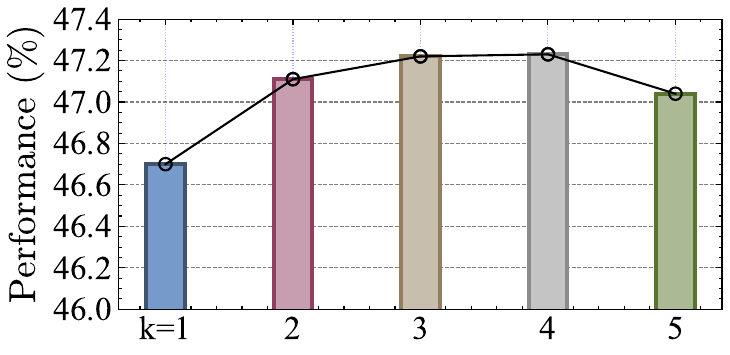}
        \caption{\small The results of experiments for hyper-parameters number of clusters.}
        \label{fig:k-value}
    \end{minipage}
    \vspace{-1.5em}
\end{figure}

\paragraph{RQ7: How do the number of intrinsic component of \model~ influence performance outcomes?}
\tian{
As Figure~\ref{fig:k-value} shown, we conduct a comprehensive and meticulous analysis by fine-tuning the Dolly-15K model on the LLaMA2-7B dataset and subsequently evaluating its performance on the MMLU benchmark to rigorously examine the impact of variations in the intrinsic component, symbolized by the variable $k$, on the model's overall performance. \textit{Empirically we find that the number k of clusters is not a sensitive parameter for \model~}, with a wide range of reasonable number k of clusters (e.g. 2 to 4) performing decently well in all settings in our experiments. Specifically, the performance loss of k=3 vs. the optimal k=4 is only 0.42\%.}
\tian{Meanwhile, as illustrated in Figure \ref{fig:k-value_gen}, we employ three distinct methods to generate the number of corpus clusters 15-fold, and the results demonstrate that the k-means \citep{k-means-1} yields comparable outcomes with DBSCAN \cite{DBSCAN}.}
Therefore, based on this observation, we choose k-means because it is simple but effective, more sophisticated hyperparameter search approaches (e.g. DBSCAN, parameter sweep and Bayesian optimization) will be unnecessarily costly. 
It's noteworthy that \model~ is adeptly designed to orchestrate its components in a way that it can automatically calibrate and navigate toward the optimal performance configuration across various parameters. This intelligent auto-tuning is achieved through the application of the k-means clustering algorithm. This strategic component orchestration not only enhances performance but also ensures a more efficient and effective utilization of resources, underpinning the model's capability to adapt and perform efficiently in a dynamic computational environment.

\section{Related work}
\paragraph{Parameter-Efficient Fine-tuning}
LLMs are becoming increasingly powerful, but fine-tuning them often requires significant computational resources. This has spurred research on parameter-efficient fine-tuning (PEFT) techniques that reduce memory and storage costs during adaptation. One prominent PEFT approach is adapters~\citep{HoulsbyGJMLGAG19,RebuffiBV17}. It introduces new, trainable dense layers within the existing model, keeping the original parameters frozen. This concept has proven successful across various domains~\citep{PfeifferKRCG21,Stickland019,Sung0B22,AutoPEFT}.  Further improvements on adapter compactness involve constructing parameter matrices using Kronecker products of low-rank matrices~\citep{MahabadiHR21}. Another PEFT strategy directly manipulates activations with learned vectors. This can be achieved through concatenation~\citep{P-tuning,LiL20, LesterAC21}, multiplication (IA$^3$;  \cite{IA3}), or addition (BitFit; \cite{ZakenGR22}). Prefix-tuning~\citep{LiL20} and prompt-tuning~\citep{LesterAC21} are noteworthy examples that fine-tune continuous prompts instead of designing discrete ones~\citep{DengWHWGSSXH22}. Interestingly, a study suggests that many PEFT methods can be viewed as a form of adapter, providing a unified perspective~\citep{HeZMBN22}. Beyond adding new parameters or altering the computational graph, researchers also explore sparse~\citep{GuoRK20,SungNR21} or low-rank updates (LoRA; \cite{HuSWALWWC22}).

\paragraph{Multi-LoRA Architecture}
LoRA has notably garnered increasing interest recently, becoming a standard approach for adapting LLMs such as LLaMA~\citep{llama,llama2} under limited computational resources. Recognizing its potential, researchers have delved deeper, exploring the benefits of employing multiple LoRAs. LoraHub~\citep{HuangLoraHub2023} takes this multi-LoRA approach by training several adapters and strategically picking combinations based on the domain during inference. Meanwhile, MultiLoRA~\citep{WangMultiLoRA2023} focuses on horizontal scaling, aiming to reduce LoRA's parameter dependence. This involves splitting LoRA modules along the rank dimension and introducing learnable scaling factors for enhanced expressiveness. Addressing scaling challenges from a different angle, the mixture of LoRA concept is further proposed~\citep{Zadouri2023}. This mitigates resource consumption when scaling instruction-tuned LLMs. Recognizing the potential for conflict during instruction tuning, LoRAMoE~\citep{DouLoRAMoE2023} leverages the Mixture-of-Experts  (MoE;~\cite{JacobsJNH91}) structure to safeguard the pre-trained LLM's knowledge from excessive corruption by instruction data. Similarly, MOELoRA~\citep{LiuMOELoRA2023} incorporates a MoE framework into LLMs, thereby improving their multitasking capabilities in the medical domain. Shifting the focus to the system perspective, S-LoRA~\citep{ShengSLoRA2023} provides a framework for efficiently serving multiple LoRA adapters. Unlike previous methods that relied on choosing LoRA combinations based on their training domains, \model~ breaks free from the dependence on domain knowledge during inference. Additionally, \model~'s asymmetric structure further enhances parameter efficiency compared to existing symmetric approaches.

\section{Conclusion}
In this work, we start by conducting exploratory experiments applying the LoRA technique to LLaMA2, aiming to adapt it to a new domain across various tasks. This study unveils the limitations of employing a single LoRA for the entire domain, highlighting the detrimental effects of domain interference. In response, we introduce a novel architecture \model~ that features an asymmetric structure with a shared matrix for all samples and distinct matrices for each intrinsic component. This design improves domain adaptation by selectively focusing on distinct components, enhancing both fine-tuning and inference efficiency. Our research highlights the importance of balancing learning capabilities for diverse datasets against the need for a lean model, offering a viable pathway for improving LLMs with minimal parameter growth. \tian{More discussion about limitation and broader impacts are available in Appendix \ref{section_limitation} and \ref{section_impacts}}.


\bibliography{main}
\bibliographystyle{plain}

\newpage
\appendix
\section{Datasets and Baselines}

\subsection{Datasets}
\label{appendix:datasets}
\paragraph{Single Domain}
\begin{enumerate}
    \item \textbf{General}: we fine-tune with the general instruction tuning databricks-dolly-15k for generic language capability and evaluate with MMLU.
\item \textbf{Medical}: we fine-tune with GenMedGPT and clinic-10k from ChatDoctor for medicine applications and evaluate medical tasks in MMLU including three related tasks: ``clinical knowledge'', ``professional medicine'' and ``college medicine''.
\item \textbf{Law}: we fine-tune with two legal instruction tuning datasets Lawyer-Instruct and US-Terms then evaluate with law tasks in MMLU including two related tasks: ``professional law'' and ``international law''.
\item \textbf{Math}: we fine-tune with the training split of GSM8K for mathematical reasoning and evaluate with test set of GSM8K.
\item \textbf{Code}: we fine-tune with CodeAlpaca for code generation and evaluate with HumanEval.
\end{enumerate}

\paragraph{Multi-task Domain} As well for complex mixed multi-task/domain, we select a portion of the Flanv2 datasets covering Natural Language Understanding (NLU) and Natural Language Generation (NLG), which can be grouped into 10 distinct task clusters. Then we evaluate it with the Big-Bench Hard (BBH) benchmark.

We summarize the details of the used datasets as follows:
\begin{enumerate}
    \item \textbf{Struct-to-Text Conversion}: This task evaluates the capability to generate natural language descriptions from structured data inputs. We use the following datasets: (1) CommonGen; (2) DART; (3) E2ENLG; (4) WebNLG;

\item \textbf{Translation}: Translation involves converting text from one language to another, maintaining the original meaning and nuances. We use the following datasets: (1) En-Fr from WMT’14; EnDe, En-Tr, En-Ru, En-Fi, En-Ro from WMT’16; (3) En-Es from Paracrawl.

\item \textbf{Commonsense Reasoning}: This involves assessing the ability to apply physical or scientific principles alongside common sense in reasoning tasks. We use the following datasets: (1) COPA, (2) HellaSwag, (3) PiQA, and (4) StoryCloze.

\item \textbf{Sentiment Analysis}: A fundamental task in natural language processing (NLP) that determines the sentiment polarity (positive or negative) of a given text. We use the following datasets: (1) IMDB, (2) Sentiment140, (3) SST-2, and (4) Yelp. information sources. We use the following datasets: (1) ARC, (2) NQ, and (3) TriviaQA.

\item \textbf{Paraphrase Detection}: This task requires models to ascertain whether two sentences convey the same meaning, indicating semantic equivalence. We use the following datasets: (1) MRPC, (2) QQP, and (3) Paws Wiki.

\item \textbf{Coreference Resolution}: Involves identifying instances within a text that refer to the same entity, demonstrating an understanding of textual context. We use the following datasets: (1) DPR and (2) WSC273.

\item \textbf{Reading comprehension}: Assesses the capability to derive answers to questions from a provided text containing relevant information. We use the following datasets: (1) BoolQ, (2) DROP, (3) MultiRC, (4) OBQA, (5) SQuADv1, (6) SQuADv2.

\item \textbf{Reading Comprehension with Commonsense}: Merges traditional reading comprehension skills with commonsense reasoning, requiring understanding beyond the explicit text. We use the following datasets: (1) CosmosQA; (2) ReCoRD.

\item \textbf{Natural Language Inference}: Focuses on deducing the relationship between two sentences, determining if the second sentence logically follows from, contradicts, or is unrelated to the first sentence. We use the following datasets: (1) ANLI, (2) CB; (3) MNLI; (4) QNLI; (5) SNLI; (6) WNLI; (7) RTE.

\item \textbf{Closed-Book Question Answering}: This task challenges models to answer questions about general knowledge without direct access to external
\end{enumerate}

\subsection{Baselines}
\label{app_baselines}
\paragraph{PEFT methods }
\begin{enumerate}
    \item \textbf{Full Fine-tuning} is the default strategy for adaptation. During fine-tuning, the model is initialized with pretrained weights and biases, and all model parameters undergo gradient updates.

\item \textbf{Prompt Tuning} adds task-specific prompts to the input, and these prompt parameters are updated independently of the pretrained model parameters which are frozen.

\item \textbf{P-Tuning} adds trainable prompt embeddings to the input that is optimized by a prompt encoder to find a better prompt, eliminating the need to manually design prompts. The prompt tokens can be added anywhere in the input sequence, and P-Tuning also introduces anchor tokens for improving performance.

\item \textbf{Prefix Tuning} prefixes a series of task-specific vectors to the input sequence that can be learned while keeping the pretrained model frozen. The prefix parameters are inserted in all of the model layers.

\item \textbf{IA3} enhances efficiency by infusing learned vectors into transformer architectures, drastically cutting trainable parameters while preserving performance and minimizing inference latency.

\item \textbf{AdaLoRA} is a method for optimizing the number of trainable parameters to assign to weight matrices and layers, unlike LoRA, which distributes parameters evenly across all modules. More parameters are budgeted for important weight matrices and layers while less important ones receive fewer parameters.
\end{enumerate}

\paragraph{Multiple LoRA weighted average methods}
\begin{enumerate}
    \item  \textbf{LoRA MoE}.
    A collection of \( n \) parameterized experts, denoted as \( E_1, ..., E_n \), is orchestrated by a router network $R$. This network features a dense layer with adjustable weights \( W_g \) from \( \mathbb{R}^{d_m \times n} \). A softmax function then processes an intermediate token representation \( x \), yielding gating scores \( s_1, ..., s_n \) that determine the weighted contribution of each expert’s output:

\begin{equation}
s_i = R(x)_i = \text{softmax}(W_g^T x) \quad (\text{Router}) 
\end{equation}

Subsequently, the overall output \( y \) is synthesized by aggregating the experts' outputs, each modulated by its respective gating score:
    \begin{equation}
    y = \sum_{i=1}^{n} s_i \cdot E_i(x) \quad (\text{MoE}) 
    \end{equation}
This results in a dynamic allocation of the model's capacity, enabling specialized processing by experts as directed by the router's gating mechanism.

    \item \textbf{LoraHub} aggregates 20 LoRAs at random for new downstream tasks. To master the weight of each LoRA, it utilizes a black-box optimization technique, bypassing the need for gradient calculations of the large model. This process involves weighted averaging at the parameter level. Mirroring the MoE training approach, we select 20 random samples for each task, creating a cohesive training dataset optimized through this black-box method.
\end{enumerate}

\section{Initialization via k-means}
\label{section: k-means}
In the case of considering heterogeneous corpora, it is crucial to select the appropriate number $N$ of matrix B, to ensure consistent performance and minimize unnecessary computational overhead. This choice is usually closely related to the training corpus. In this work, we propose initializing \model~ modules via $k$-means~\cite{k-means-1} algorithm for adaptive initialization. Specifically, $k$-means is utilized to process the heterogeneous corpus to identify the best-fit taxonomy of the corpus, i.e., the optimal $k$. First, we extract key features from the corpus by applying the Term Frequency-Inverse Document Frequency (TF-IDF;~\cite{TF-IDF}) algorithm and transform the textual information into numerical feature vectors. We integrate the elbow method~\cite{elbow} to determine the optimal value of $K$. Initially, $K$ cluster centers are randomly selected for preliminary clustering as Eq.~\ref{eq: cluster_1}, followed by updating the cluster centers to accurately reflect the data within each cluster as Eq.~\ref{eq: cluster_2}. where $C_j$ is the cluster center to which data point $X_{i}$ is assigned and $d(\cdot , \cdot)$ is the Euclidean distance function. $S_j$ is the set of data points in the $j$-th cluster. 
\begin{align}
\label{eq: cluster_1}
C_j &= \underset{C_j}{\mathrm{argmin}}\ d(X_i, C_j) \\
C_j &= \frac{1}{|S_j|} \sum_{X_i \in S_j} X_i
\label{eq: cluster_2}
\end{align}

By analyzing the relationship between the sum of squares of errors (SSE) and different $K$ values, we observe that SSE decreases as $K$ increases. Identifying the elbow point on the SSE curve --- where the rate of decrease in SSE slows down  --- is crucial. The elbow point represents the optimal $K$ value, beyond which increasing the number of clusters does not significantly enhance performance, thereby achieving an ideal balance between model complexity and performance.

\section{LoRA Breakdown}
\label{app:lora}

\begin{figure}[!ht]
    \centering
    \small
    \subfigure[Compare fine-tuned LoRA modules of GSM8K~\cite{gsm8k} with its subsets using T-SNE. We employ the Independent and Identically Distributed (IID) segmentation scheme to divide GSM8K into three subsets and fine-tune them using different LoRAs.]{
    \includegraphics[width =0.33\linewidth]{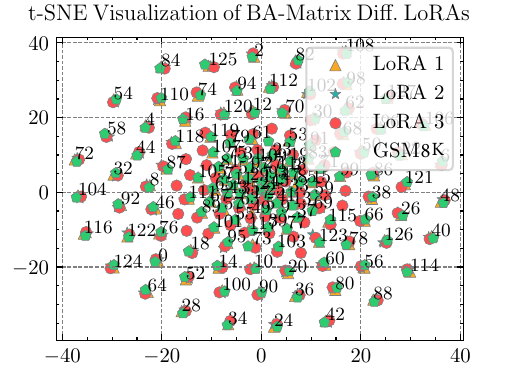}
    \includegraphics[width=0.325\linewidth]{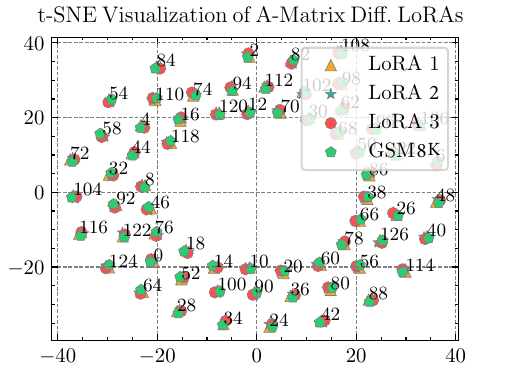}
    \includegraphics[width=0.325\linewidth]{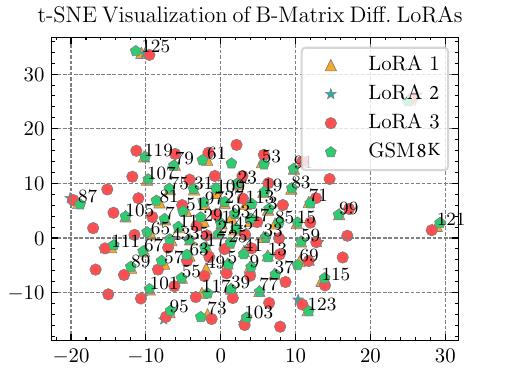}}
    \subfigure[Specific tasks.]{
    \includegraphics[width =0.33\linewidth]{figures/LoRA_weight_layer_BA_42.pdf}
    \includegraphics[width=0.325\linewidth]{figures/LoRA_weight_layer_A_42.pdf}
    \includegraphics[width=0.325\linewidth]{figures/LoRA_weight_layer_B_42.pdf} }
    \caption{Breakdown analysis of LoRA modules. Compare fine-tuned LoRA modules of GSM-8K~\cite{gsm8k} with its subsets using T-SNE. We employ the Independent and Identically Distributed (IID) segmentation scheme to divide GSM8K into three subsets and fine-tune them using different LoRAs. Consider LLaMA2-7B (random seed=42), which contains 32 decoder layers, corresponding to 32 adaptive modules. Each module consists of \{0: q\_proj\_A, 1: q\_proj\_B, 2: v\_proj\_A, 3: v\_proj\_B\} submodules. This makes a total of $32\times4$ submodules. (a,b) left displays all submodules. (a,b) center shows all even submodules, i.e. the A-matrix. (a,b) right represents all odd submodules, i.e. the B-matrix. It can be seen that the differences in the fine-tuned LoRA modules for different tasks arise mainly from the B matrix.}
\end{figure}

\begin{figure}[!t]
    \centering
    \includegraphics[width=0.4\linewidth]{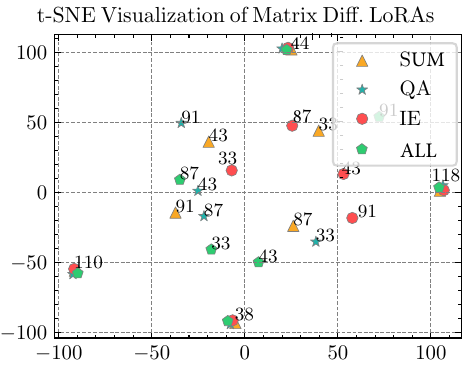}
    \caption{Breakdown analysis of LoRA modules (Dolly-15K) with its subsets using T-SNE on different layer.}
\end{figure}

\section{More Results}
Tabel ~\ref{table:mix-task-corpus_app} shows Comparative performance of different tuning schemes, including base model (Base), LoRA tuning (LoRA), LoraHub learning, multi-LoRA tuning with MoE inference (LoRA MoE) and our proposed \model~ learning across mix-task domain on the BBH benchmark with LLaMA2-7B as the base LLM (3-shot).
\label{app:results}
\begin{table}[!t]
    \centering
    \caption{Comparative performance of different tuning schemes, including base model (Base), LoRA tuning (LoRA), LoraHub learning, multi-LoRA tuning with MoE inference (LoRA MoE) and our proposed \model~ learning across mix-task domain on the BBH benchmark with LLaMA2-7B as the base LLM (3-shot).}
    \resizebox{0.95\linewidth}{!}{
\begin{tabular}{c|cccc>{\columncolor{blue!20}}c}
\toprule[1.5pt]
Task  & Base  & LoRA  & LoraHub   & LoRA MoE &   \model~ \\
\midrule
Boolean Expressions & 61.9 & 67.1 & 72.9 & 68.0 & 73.7 \\
Causal Judgement & 52.2 & 54.9 & 50.1 & 51.4 & 53.2 \\
Date Understanding & 30.4 & 35.2 & 36.0 & 33.9 & 36.0 \\
Disambiguation & 34.8 & 45.2 & 49.1 & 47.2 & 50.3 \\ 
Dyck Languages  & 15.8 & 18.7 & 14.5 & 16.8 & 19.8 \\
Formal Fallacies  & 49.0 & 62.2 & 64.5 & 67.6 & 65.3 \\
Geometric Shapes  & 9.7 & 17.7 & 18.7 & 17.7 & 19.7 \\
Hyperbaton  & 51.8 & 74.3 & 74.3 & 68.9 & 77.2 \\
Logical Deduction (five objects)  & 21.9 & 33.3 & 38.7 & 40.0 & 42.2 \\
Logical Deduction (seven objects) & 15.0 & 36.4 & 37.3 & 40.7 & 40.7 \\
Logical Deduction (three objects) & 32.8 & 41.4 & 38.5 & 43.7 & 42.9 \\
Movie Recommendation  & 34.4 & 53.5 & 56.0 & 56.8 & 58.3 \\
Multistep Arithmetic & 1.2 & 1.2 & 1.9 & 1.9 & 1.8 \\
Navigate & 53.8 & 52.7 & 56.2 & 58.0 & 57.1 \\
Object Counting& 40.1 & 40.5 & 42.3 & 44.7 & 42.3 \\
Penguins in a Table& 21.7 & 23.2 & 25.0 & 23.2 & 25.9 \\
Reasoning about Colored Objects& 19.4 & 28.0 & 32.7 & 38.3 & 38.3 \\
Ruin Names& 24.3 & 28.7 & 34.3 & 34.3 & 36.7 \\
Salient Translation Error Detection& 11.3 & 11.1 & 17.1 & 16.2 & 20.1 \\
Snarks& 44.0 & 47.9 & 54.9 & 53.6 & 56.9 \\
Sports Understanding& 57.5 & 59.0 & 61.2 & 59.0 & 60.2 \\
Temporal Sequences& 21.1 & 32.6 & 28.9 & 34.1 & 30.4 \\
Tracking Shuffled Objects (five objects)& 21.9 & 23.7 & 23.7 &  28.0 & 29.3 \\
Tracking Shuffled Objects (seven objects)& 14.6 & 15.3 & 16.6 & 15.3 & 15.3 \\
Tracking Shuffled Objects (three objects)& 32.4 & 38.4 & 39.0 & 38.4 & 40.7 \\
Web of Lies& 51.4 & 52.8 & 53.2 & 50.1 & 52.0 \\
Word Sorting & 29.6 & 33.6 & 33.6 &  31.2& 34.0 \\
\midrule[0.75pt]
 Avg Performance& 31.6 & 36.8 & 39.7 & 40.3 & 41.5 \\
\# of A/B for training & 0/0 & 1/1 & 48/48 & 48/48 & 1/10 \\
\# of A/B for inference & 0/0 & 1/1 & 20/20 & 48/48 & 1/10 \\
\% Params & - & 0.062 &  1.240 &  2.976 & 0.341 \\
\bottomrule[1.5pt]
\end{tabular}}
\label{table:mix-task-corpus_app}
\end{table}

\section{Limitation}
\label{section_limitation}
\tian{
\model~ is computationally demanding, primarily due to the necessity of fine-tuning large-scale language models. It incurs a higher training expenditure than conventional PEFT methods, attributed to the employment of multiple adapter copies. Empirical data suggest that \model~ requires 1 to 2 times more training iterations compared to typical PEFT methods, which adversely affects the environmental footprint of model training. The \model~ framework, distinct from prevailing PEFT approaches, holds promise for enhancing the efficacy of existing PEFT strategies. In our current study, we examine established PEFT techniques --- LoRA. However, we have not tested additional configurations such as prompt-tuning and adapter, deferring these explorations to subsequent research. Additionally, our assessment is exclusively within the context of fine-tuning. Exploration of its efficacy during the pre-training phase remains an avenue for future research.}

\section{Broader Impacts}
\label{section_impacts}

\paragraph{Positive Societal Impacts}
The proposed \model~ framework, with its asymmetric structure and parameter-efficient fine-tuning approach, has the potential to make LLMs more accessible and efficient. This could democratize AI, enabling more researchers, developers, and organizations to leverage the power of LLMs for various applications, ultimately driving innovation and progress. Moreover, by effectively addressing the challenge of domain or task interference, \model~ could significantly enhance the performance of LLMs in complex, multi-task domains. This could lead to more accurate and reliable AI-powered tools and services in areas like healthcare, education, and finance, ultimately improving the quality of life for many people. Lastly, the parameter-efficient approach of \model~ could help reduce the computational resources required for training and fine-tuning LLMs, thereby lessening the environmental impact of AI.

\paragraph{Negative Societal Impacts}
As with any AI technology, there are potential negative societal impacts to consider. The risk of misuse is a significant concern, as \model~ could be used for malicious purposes, such as creating more sophisticated and convincing deepfakes or spreading misinformation and propaganda. Additionally, the increased efficiency and accessibility of AI brought about by \model~ could lead to job displacement in certain sectors, as AI-powered tools and services become capable of performing tasks traditionally done by humans. Lastly, the use of LLMs in various applications could potentially lead to privacy and security issues, especially if these models are used to process or generate sensitive information. The proposed \model~ framework, while not directly related to these issues, could inadvertently contribute to them by making it easier to deploy LLMs in various applications.


\newpage

\end{document}